\documentclass[
]{ceurart}

\usepackage{listings}
\usepackage{algorithm}
\usepackage{algorithmic}
\usepackage{pifont}
\newcommand{\xmark}{\ding{55}}
\begin{document}

%%
%% Rights management information.
%% CC-BY is default license.
\copyrightyear{2023}
\copyrightclause{Copyright for this paper by its authors.Use permitted under Creative Commons License Attribution 4.0
  International (CC BY 4.0).}

%%
%% This command is for the conference information
\conference{AIRO 2023 the 10th Italian Workshop on Artificial Intelligence and Robotics co-located with the 22nd International Conference of the Italian Association for Artificial Intelligence (AI*IA 2023), Rome, Italy}

%%
%% The "title" command
%\title{Scalable \#DNN-Verification Through Parallel Computing}

%\title{A Parallel Computational Approach for Accelerating \#DNN-Verification}

%\title{Enhancing Efficiency and Scalability of \#DNN-Verification via Parallel Computation}

\title{Scaling \#DNN-Verification Tools with\\ Efficient Bound Propagation and Parallel Computing}

%%
%% The "author" command and its associated commands are used to define
%% the authors and their affiliations.
\author[]{Luca Marzari}[%
email=luca.marzari@univr.it]
\cormark[1]

\author[]{Gabriele Roncolato}[%
email=gabriele.roncolato@studenti.univr.it]

\author[]{Alessandro Farinelli}[%
email=alessandro.farinelli@univr.it]

\address[]{Department of Computer Science, Univerisity of Verona, Italy}

%% Footnotes
\cortext[1]{Corresponding author.}
%\fntext[1]{These authors contributed equally.}

%%
%% The abstract is a short summary of the work to be presented in the
%% article.
\begin{abstract}
  Deep Neural Networks (DNNs) are powerful tools that have shown extraordinary results in many scenarios, ranging from pattern recognition to complex robotic problems. However, their intricate designs and lack of transparency raise safety concerns when applied in real-world applications. In this context, Formal Verification (FV) of DNNs has emerged as a valuable solution to provide provable guarantees on the safety aspect. Nonetheless, the binary answer (i.e., \texttt{safe} or \texttt{unsafe}) could be not informative enough for direct safety interventions such as safety model ranking or selection. To address this limitation, the FV problem has recently been extended to the counting version, called \textit{\#DNN-Verification}, for the computation of the size of the unsafe regions in a given safety property's domain. 
  Still, due to the complexity of the problem, existing solutions struggle to scale on real-world robotic scenarios, where the DNN can be large and complex. To address this limitation, inspired by advances in FV, in this work, we propose a novel strategy based on reachability analysis combined with Symbolic Linear Relaxation and parallel computing to enhance the efficiency of existing exact and approximate FV for DNN counters. The empirical evaluation on standard FV benchmarks and realistic robotic scenarios shows a remarkable improvement in scalability and efficiency, enabling the use of such techniques even for complex robotic applications.
\end{abstract}

%%
%% Keywords. The author(s) should pick words that accurately describe
%% the work being presented. Separate the keywords with commas.

%%
%% This command processes the author and affiliation and title
%% information and builds the first part of the formatted document.
\maketitle

\section{Introduction}\label{intro}

\begin{figure}[h!]
    \centering
    \includegraphics[width=0.9\linewidth]{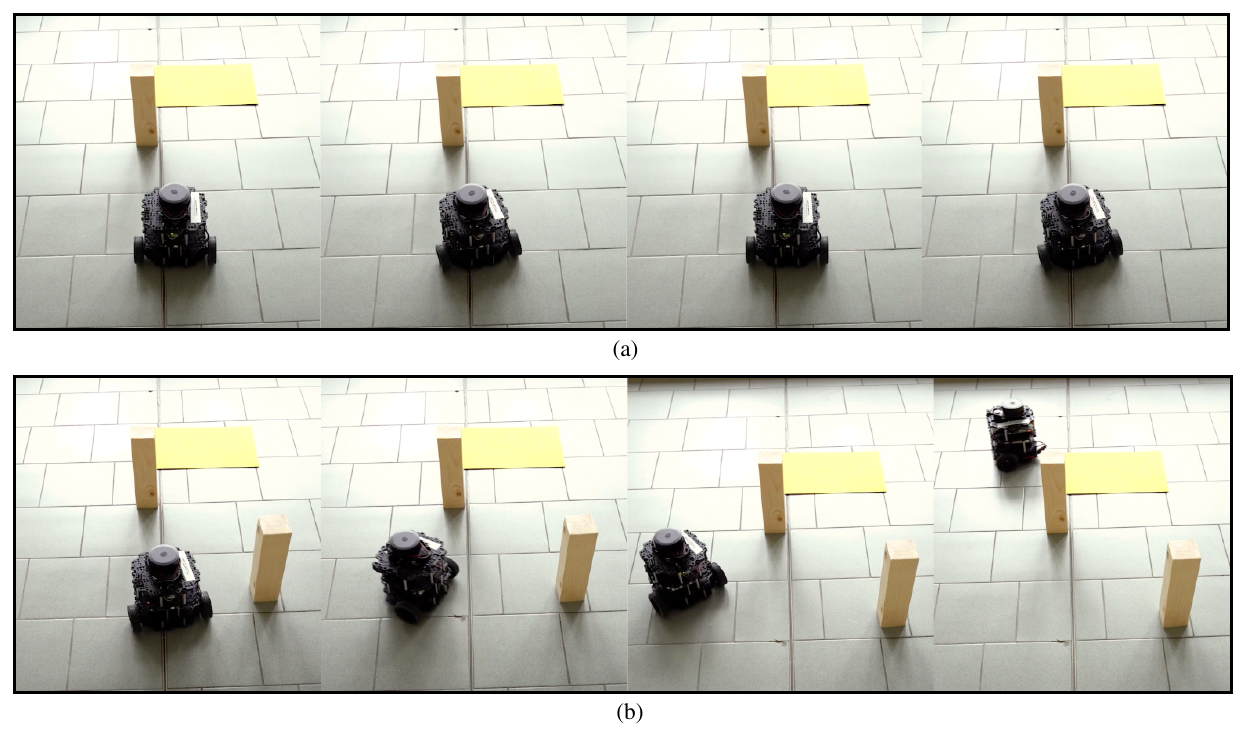}
    \caption{Explanatory image of adversarial input in a DRL setup. The robot is trained and is generally able to navigate and reach the yellow target in the environment. FV detects this unsafe input configuration where the agent shows a suboptimal behavior as it is stuck in an infinite alternating loop (a). When an obstacle is added, changing the agent's observation, the robot is able to escape from the loop and reach the target (b).}
    \label{fig:adversarial}
\end{figure}

In recent years, the use of Deep Neural Networks (DNNs) has increased due to their ability to learn complex patterns from vast amounts of data. In detail, DNNs have emerged as a novel technology revolutionizing various fields ranging from image classification \cite{image}, robotic manipulation \cite{manipulation1, manipulation2}, robotics for medical applications \cite{colon, tissue}, and even for autonomous navigation \cite{Curriculum, navigation}. As the DNNs become more powerful and pervasive in our applications, the safety aspect has become increasingly prominent. The typical non-linear and non-convex nature of these approximator functions raises two critical concerns:
(i) the behavior of these networks is often not comprehensible, leading them to be called "black boxes" (ii) slight human-imperceptible changes in the domain of these functions can cause drastic mispredictions that can endanger both the safety aspect of the intelligent agents and the human life in the real-world applications. More specifically, DNNs are vulnerable to the so-called "adversarial attacks"\cite{adversarial}. To provide the reader with an intuition of what these adversarial inputs are in practice in a robotic scenario, suppose to consider a context where we have a mobile robot trained to navigate in a particular environment. The agent's goal is to navigate towards a random target placed in the environment without having access to a map, using only Lidar values to sense the surroundings. In this context, if we train a neural network, for instance, with some Deep Reinforcement Learning (DRL) techniques, it has been shown in \cite{TACAS} how, despite empirically achieving a high level of success rate (measured in terms of how many times the agent reaches the goal) and safety (measured in terms of collision rate with obstacles), it is possible to find particular input configurations, that when propagated through the network, lead to suboptimal behaviors, such as the one shown in Fig. \ref{fig:adversarial}.

To address such an issue, the research field of Formal Verification (FV) of DNNs \cite{Liu}, has emerged as a valuable solution to provide formal assurances on the safety aspect of these functions before the actual deployment in real scenarios. Broadly speaking, the goal of FV is to prove (or falsify) a desired input-output relationship--aka safety property-- for a specific DNN. This decision problem is known as \textit{DNN-Verification} and is typically solved either using interval analysis \cite{Moore} to propagate the input bound through the network and perform a \textit{reachability analysis} in the output layer \cite{reluval,Neurify,BetaCrown} or by encoding the linear combinations and the non-linear activation functions of a DNN as constrained for an optimization problem \cite{Reluplex, Marabou}. 

Despite the considerable advancements made by DNN-verifiers over the years \cite{Liu}, the binary nature (\texttt{safe} or \texttt{unsafe}) of the result provided by these tools could hide additional helpful information to gain a deeper understanding of these functions. In particular, with the \textit{DNN-Verification}, we are able to find a counterexample in case a particular safety property does not hold, however, we do not have any information on how many unsafe configurations violate the safety property for the given DNN. This is a crucial element to rank different trained models or to re-train the model so to avoid such unsafe states. To this end, in \cite{Baluta}, the authors proposed a quantitative variant of FV, aiming to provide the total number of violations for given safety property and a specific category of DNNs called Binarized Neural Networks (BNNs). Nonetheless, a potential binarization of a DNN to a BNN typically does not preserve violation points \cite{ZhangBaluta}, hence, \cite{CountingProVe} formalized the problem for standard DNNs called \textit{\#DNN-Verification}. This novel type of FV allows us to estimate the total number of violations in the domain of the safety property and the probability that a DNN violates a given property. However, given the NP-Completness of the standard \textit{DNN-Verification}, the \textit{\#DNN-Verification} has been proven to be \#P-Complete \cite{CountingProVe}. Hence, to solve this challenging problem, both studies in  \cite{Baluta,CountingProVe}, focus on efficient approximate solutions providing provable (probabilistic) guarantees regarding the computed count. 
The main solutions of FV tools at the state-of-the-art for both \textit{DNN-Verification} and \textit{\#DNN-Verification} are strongly based on the Branch-And-Bound (BaB) approach \cite{BaB}. However, unlike the methods to solve the \textit{DNN-Verification} problem, which potentially explore only part of the tree generated through BaB, since they seek for a single counterexample, to solve in a sound and complete fashion the \textit{\#DNN-Verification} problem, we have to explore every single node, which exponentially increases the complexity, resulting in a prohibitive computational demand.

Hence, inspired by the recent BaB-based solutions employed to solve efficiently the \textit{DNN-Verification} \cite{BetaCrown, verinet}, in this work, we first investigate possible optimizations to solve the \textit{\#DNN-Verification} in an exact fashion \cite{ProVe}. Moreover, we employ these optimizations to the recent approximate solution \cite{CountingProVe} to assess the scalability and efficiency improvements. We argue that by integrating similar optimization of standard FV tools on previously proposed exact and approximate solutions for the \textit{\#DNN-Verification} problem, we are able to enhance the scalability and efficiency of these solutions. More specifically, our intuitions rely on the fact that every time we develop a new level of the BaB tree, instead of computing the result on each node iteratively, we can employ the power computation of GPUs to check multiple scenarios simultaneously. In addition, in order to maintain an optimal branching tree size, we can employ in our optimization the \textit{Symbolic Linear Relaxation} (SLR) \cite{Neurify, verinet}, one of the most recent and efficient techniques to compute a tight bound propagation through the DNN.

The empirical evaluation on standard FV benchmarks as ACAS xu \cite{ACAS} and on realistic robotic DRL mapless navigation domain confirm our hypothesis. In particular, we empirically confirm the improvement in scalability even in complex robotic scenarios, where previous approaches for the \textit{\#DNN-Verification} problem struggled to scale, making this new type of verification applicable also in different robotic real-world applications. 

Summarizing, the main contributions of this work are the following:
\begin{itemize}
    \item we proposed an efficient bound propagation method based on BaB and SLR for solving the \textit{\#DNN-Verification} efficiently.
    \item inspired by existing FV tools that exploit GPU to speed up the verification process \cite{BetaCrown, verinet, ProVe} we employ the parallel computation of each BaB layer to enhance the performance of existing exact count approaches. 
    \item we empirically show the effectiveness and the scalability improvement of the solution proposed on standard FV benchmark (ACAS xu) and realistic robotic scenarios. 
\end{itemize}

Crucially, this work paves the way for the application of different types of FV even for complex DNNs typically employed in challenging robotic scenarios.

%\vspace{-0.6cm}
\section{Background}\label{background}
Our work exploits two main components common to most recent state-of-the-art verifiers: interval analysis with BaB approach that integrates symbolic linear relaxation to compute tight output reachable sets and parallel computation \cite{Neurify, verinet, BetaCrown}. In the following sections, we briefly introduce the \textit{DNN-Verification} and the strategies to solve the problem efficiently. Finally, we provide the formulation on the \textit{\#DNN-Verification} and the existing solutions to this quantitative variant of standard FV.  We refer interested readers to \cite{Neurify, verinet, BetaCrown, ProVe, CountingProVe} for more details.
\subsection{DNN-Verification}\label{DNN-ver}
The \textit{DNN-Verification} can be defined in two main ways, either using the satisfiability formulation or the reachability one. In general, the \textit{DNN-Verification} problem takes as input a tuple $\mathcal{T} = \langle \mathcal{N}, \mathcal{P}, \mathcal{Q}\rangle$, where $\mathcal{N}$ is a trained DNN, while $\langle \mathcal{P}, \mathcal{Q}\rangle$ encodes respectively the safety property as input-output relationship. More specifically, $\mathcal{P}$ is a precondition on the input that defines the possible input configuration we are interested in (typically expressed using the cartesian product of hyperrectangles), while $\mathcal{Q}$ encompasses the postcondition that represents the output results we aim to guarantee formally. Hence, the problem consists of verifying that for each input that satisfies $\mathcal{P}$  when fed to the DNN $\mathcal{N}$, the resulting output also satisfies $\mathcal{Q}$ \cite{Reluplex, Liu, CountingProVe}.

Referring to Fig. \ref{fig:adversarial}, assuming an observation space composed of normalized laser scan values in $[0,1]$ (where 0 represents the closest distance to an obstacle), a possible high-level "behavioral" safety property in this specific example could be:
\begin{align*}
    \mathcal{P}&: \text{frontal lidar scans } \in [0.01, 0.05]\\
    \mathcal{Q}&: \text{action } \neq \;\uparrow (\text{forward movement})
\end{align*}

Typically, what is done by the FV tools is to search for a single counterexample that falsifies the property. Hence, the FV tool searches for the existence of at least a single input configuration that satisfies $\mathcal{P}$  and the negation of the postcondition $\mathcal{Q}$. Referring to the safety property reported above, a FV tool searches for the existence of and input $x\in [0.01, 0.05]$ for which $\mathcal{N}(x) =\; \uparrow$. If this particular input configuration exists, then the FV returns a \texttt{SAT} answer, meaning that there exists at least one violation point, and therefore, the original property does not hold, i.e., the DNN is \texttt{unsafe}. Otherwise, the answer is \texttt{UNSAT}, so the original property holds for each possible input selected from the intervals encoded in $\mathcal{P}$, and the DNN is provable \texttt{safe}.

\begin{figure}[b]
    \vspace{-2mm}
    \centering
    \includegraphics[width=0.8\linewidth]{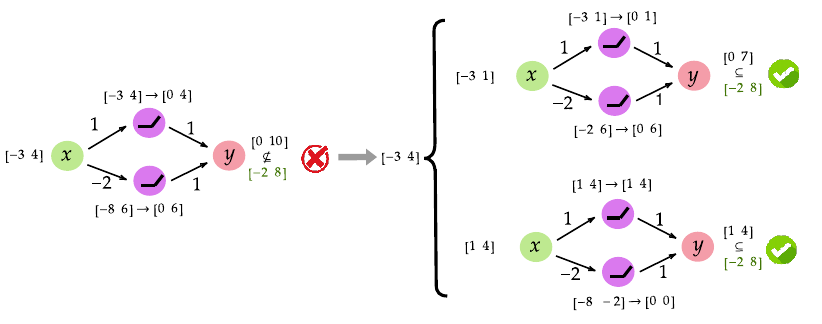}
    \caption{Explanatory image of reachability analysis combined with branching method called \textit{iterative refinement}\cite{reluval}. Exploiting the interval algebra \cite{Moore}, the lower and the upper bounds of each interval are propagated through the DNN layer-by-layer. In the left part, naive interval analysis produces a very large overestimation of the reachable set. Using input split refinements (right part) we are able to reduce the over-approximation and provably verify the property.}
    \label{fig:iterative}
\end{figure}

As previously mentioned, our approach is based on interval analysis. In this context, FV methods propagate directly the intervals that encode the precondition and perform layer-by-layer reachability analysis to compute the output reachable set $\mathcal{R}(\mathcal{X}, \mathcal{N})$, with $\mathcal{X}$ the intervals that encode the property precondition $\mathcal{P}$, and $\mathcal{N}$, our DNN. At the end of this propagation, the tool checks whether $\mathcal{R}(\mathcal{X}, \mathcal{N}) \subseteq \mathcal{Y}$, where $\mathcal{Y}$ is the desired reachable set. However, given the non-linear and non-convex nature of DNNs, estimating the exact $\mathcal{R}(\mathcal{X}, \mathcal{N})$ is infeasible, and thus, the output reachable set is typically over-approximated. Nonetheless, the naive interval analysis produces large overestimation errors as it ignores the input dependencies during interval propagation \cite{Moore}. To address this issue, \cite{reluval} proposes (i) the iterative refinement process and (ii) symbolic interval propagation. Usually, FV methods that exploit these solutions are combined with branching  \cite{BaB} methods and are called \textit{search-reachability} methods \cite{Liu}. In the next section, we briefly introduce the intuition behind these optimizations.

\subsection{Efficient Bound Estimation via Symbolic Interval Propagation and Linear Relaxation}
In Fig. \ref{fig:iterative}, we report a representation of the reachability analysis combined with a branching method called \textit{iterative refinement}. In particular, the idea is to exploit the fact that any neural network with a finite number of layers is Lipschitz continuous. Hence, leveraging the fact that the dependency error for Lipschitz continuous functions decreases as the width of intervals decreases, the iterative refinement process allows us to bisect the input interval by evenly dividing the interval into the union of two consecutive sub-intervals obtaining more tight reachable sets w.r.t. the naive propagation \cite{reluval}.

\begin{figure}[h!]
    \vspace{-2mm}
    \centering
    \includegraphics[width=\linewidth]{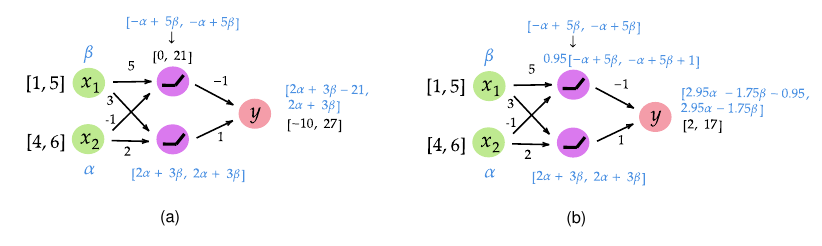}
    \vspace{-3mm}
    \caption{Explanatory image of (a) Symbolic interval propagation proposed in \cite{reluval} and (b) Symbolic linear relaxation \cite{Neurify}.}
    \label{fig:SIP-SLR}
\end{figure}

However, another source of over-approximation is given by the fact that naive interval propagation--even when combined with iterative refinement-- does not take into account all the inter-dependencies with the previous layers. To this end, \cite{reluval} proposes the Symbolic Interval Propagation (SIP) as a novel solution to preserve as much dependency information as possible while propagating the bounds through the DNNs layers. In particular, the authors consider DNNs with ReLU activation functions, and they seek to maintain linear equations for all the propagation, simplifying the non-linear nature of the neural networks and significantly speeding up the verification process. Nonetheless, ReLU nodes can demonstrate non-linear behavior for a given input interval. More specifically, if the interval in input to the ReLU node produces a result with a negative lower bound and a positive upper bound, a concretization process is required, thus losing all the inter-dependences preserved up to that specific node. To address such an issue, \cite{Neurify} proposed the Symbolic Linear Relaxation of ReLU nodes, which allows in these scenarios to partially keep the input dependencies during interval propagation by maintaining symbolic equations in the form:
\begin{equation}\label{SLR}
    z = \Big[\frac{u}{u-\ell}Eq_\ell,\; \frac{u}{u-\ell}(Eq_u-\ell)\Big]
\end{equation}

where $z$ is the ReLU node, and $\ell, u$ denote the concrete lower and upper bounds for $Eq_\ell$ and $Eq_u$, respectively. More specifically, in Fig. \ref{fig:SIP-SLR}, we report an explanatory example of interval propagation using SIP and SLR. Focusing on the left part of the image and considering the upper ReLU node, we have to perform a concretization since, considering the input bounds, we have a possible negative lower and a positive upper bound. This process leads to the loss of all the input dependencies preserved until that moment. In contrast, on the right part of the figure, using Eq. \ref{SLR} on the same ReLU node, we are able to partially preserve the information and thus obtain a tighter output reachable set.

\subsection{\#DNN-Verification}\label{background_sharpDNN}

The combination of the optimizations seen in the previous sections allows formal verification tools to achieve impressive performance, enabling the ability to verify several safety properties in a fraction of a second \cite{VNNcomp}. Nonetheless, one of the main limitations of FV lies in the binary nature of the result returned. To address this issue, in \cite{ProVe}, the authors propose \texttt{ProVe} a novel FV method to compute the so-called  \textit{Violation Rate} (VR), a metric to infer the percentage of the property's input area that causes a violation of a given safety specification. Moreover, in a recent work, \cite{CountingProVe} formalized the \textit{\#DNN-Verification}, the counting version of the \textit{DNN-Verification} and proposed, due the \#P-completness of the problem, an efficient approximate solution called \texttt{CountingProVe}. More specifically, this type of problem seeks to count the provable number of points (or regions) that satisfy (or not) a given safety property allowing more safety-impactful operations such as estimating the probability of incurring unsafe behaviors, ranking the model for its safety level and subsequent selection of the safest one, or even guiding the training of a DRL agent in a more informed fashion, for instance, minimizing this number over the training process \cite{aamas_safety, CROP}.

To solve this problem in a sound and complete fashion, the idea is to formally verify each node of the Branch-and-Bound \cite{BaB} tree iteratively. Each node of this BaB represents a partition of the property's domain where, for some part of the domain, the property holds and for another one, it does not. Hence, in the leaves of the BaB tree, we have either a situation of a completely safe or unsafe region. We are then able to obtain the total count of violations, computing the total volume of the subinput spaces in the leaves that present complete violations. However, to state whether each node of the BaB is safe or not, there is the necessity of solving a \textit{DNN-Verification} instance, and thus, this approach struggles to scale on real-world scenarios. To address this issue, \cite{CountingProVe} proposes a novel approximate solution called \texttt{CountingProVe} providing provable (probabilistic) guarantees on the count returned. The underlying idea of this approximation is based on maintaining a well-balanced distribution of violation points using a sampling approach during each splitting operation. At the conclusion of this process, the goal is to evaluate only the count within a single leaf, multiplying this count by a factor of $2^s$, where $s$ denotes the number of splits performed. However, achieving a perfect balance among these violation points within the BaB tree is unattainable, prompting the authors to propose a probabilistic solution in expectation and calling an exact count only at the end of this process. This approach is more efficient compared to an exact method, as it requires a single invocation of the exact count method on a reduced problem instance. 

Nevertheless, the probabilistic nature of point balancing and the requirement for an exact counting method may still present challenges in contrast to recent approximate solutions \cite{aamas_safety, eProve}, which provide slightly less precise bounds on the estimated violation rate, but significantly reduce the computation time by orders of magnitude. In particular, in order to estimate the number of states where a binary property holds (i.e., whether the agent commits a violation or not), in these approaches, a sampling of a small subset of states from a potentially infinite one is used, showing that the result obtained from this subset is representative of the original set we are interested in. The theoretical guarantees of these approaches lie either directly from the Chernoff bound \cite{mitzenmacher2005probability} or on the statistical prediction of tolerance limits of \cite{wilks1942statistical}. Even though these approaches allow us to obtain a result in significantly less computation time, with respect to the methods proposed in this work, they could fail to capture minimal percentage values of the violation rate (as shown in our experiments) and thus could not always be applicable in realistic safety-critical robotic domains. 

\section{Efficient \#DNN-Verification via Symbolic Linear Relaxation and Parallel Computing}\label{method}
In this section, we present our main contributions. Considering the fact that the \#DNN-Verification problem is \#P-complete, the time complexity of the algorithm for exact counting the violations in the property's domain deteriorates rapidly as the instance input size of the network grows. In fact, even moderately small networks with only five input nodes become unmanageable using this approach. In particular, what makes the problem so complex is having to solve an instance of the \textit{DNN-Verification} problem on each node of the BaB. As the size of the input increases, we have a larger number of potential splits to perform, significantly increasing the complexity of the problem. In fact, suppose to consider a network of only $5$ input nodes, where on each dimension, we perform $12$ splits to be able to get an exact count of the \textit{violation rate}, thus getting $5\cdot2^{12}$ nodes to verify. Assuming that each node of the BaB requires an average time of $5$ seconds to get an answer, it would require about $5\cdot2^{12}\cdot5s = 102400s \sim 28$ hours in order to compute the total count of violations in the property's domain.  

To address this challenging problem, inspired by the existing optimizations of \textit{search-reachability} methods for the \textit{DNN-Verification} problem, we propose to combine the symbolic linear relaxation \cite{Neurify} to reduce the number of splits required during the computation, and the parallel computation to make the \textit{\#DNN-Verification} problem manageable. We implement our exact count for ReLU DNNs, even though our approach is easily extendable to different DNNs such as Tanh or Sigmoid, using similar solutions proposed in \cite{verinet} to deal with different non-linear activation functions.

\begin{algorithm}[t]
\caption{\texttt{Exact Count}}\label{alg:exact}
\begin{algorithmic}[1]
\small
\STATE \textbf{Input:} $\mathcal{T} =\langle\mathcal{N},\mathcal{P}, \mathcal{Q}\rangle$
\STATE \textbf{Output: } Violation Rate (VR)
\vspace{0.2cm}

\STATE  $VR$ $\gets 0,$ unknown $\gets \mathcal{P}$ 
\WHILE{unknown $\neq \emptyset$}
    \STATE  $\mathcal{X} \gets$ \texttt{Pop}(unknown) 
    \STATE $\mathcal{R} \gets$ \texttt{ComputeReachableSet}($\mathcal{X}$, $\mathcal{N}$)
    \IF {$\mathcal{R} \subseteq \mathcal{Y}$}
        \STATE $VR\gets VR+ \vert \mathcal{R} \vert$
    \ELSIF{$\mathcal{R} \cap \mathcal{Y} = \emptyset$}
        \STATE unknown $\gets $ unknown $ \setminus\; \mathcal{X}$ 
    \ELSE
        \STATE unknown $\gets $ unknown $ \cup$ \texttt{IterativeRefinement}(Area)
    \ENDIF
    
\ENDWHILE
%\ENDIF
\STATE \textbf{return} \textit{VR}
\end{algorithmic}
\end{algorithm}

 We report in Alg. \ref{alg:exact} the complete pipeline of our exact count method. We start by considering the entire property's domain $\mathcal{X}$ which encodes the property's precondition, and we check for the negated property's postcondition $\mathcal{Q}$ (as discussed in Sec.\ref{DNN-ver}). Broadly speaking, we check if the corresponding output reachable set of $\mathcal{X}$ lies in the undesired reachable set $\mathcal{Y}$. For each node of each hidden layer, we store two linear equations, one for the lower and one for the upper bound of the interval. Once we reach a ReLU node, in case concretization is needed, we use the symbolic linear relaxation, i.e., Eq. \ref{SLR}, described in Sec. \ref{background}, to preserve the interdependence with the previous layer, obtaining tight output bounds and thus speeding up the verification process. In the output layer, we check whether the output reachable set $\mathcal{R}(\mathcal{X}, \mathcal{N})$ is contained or not in $\mathcal{Y}$--the output reachable set that encodes the negated postcondition $\mathcal{Q}$-- using Moore's interval algebra \cite{Moore} and, in particular, this relation:

\[\mathcal{R}(\mathcal{X}, \mathcal{N}) \subseteq \mathcal{Y} \iff (\mathcal{R}_\ell \geq \mathcal{Y}_\ell \wedge \mathcal{R}_u \leq \mathcal{Y}_u)\]

where $\mathcal{R}_\ell$, ,$\mathcal{Y}_\ell$ are the lower bound and $\mathcal{R}_u$, $\mathcal{Y}_u$ are the upper bound of the reachable sets $\mathcal{R}(\mathcal{X}, \mathcal{N})$ and $\mathcal{Y}$, respectively. 

Since we start verifying the negation of the postcondition, if the $\mathcal{R}(\mathcal{X}, \mathcal{N}) \subseteq \mathcal{Y}$ this implies that all $\mathcal{X}$ is unsafe, that is, there exists at least one and possibly an infinite number of violation points\footnote{Since we operate with a continuous input space hence we could have an infinite number of violations.}. On the other hand, if 
$\mathcal{R}(\mathcal{X}, \mathcal{N}) \cap \mathcal{Y} = \emptyset$, we can state that there are no violation points in $\mathcal{X}$, thus the original safety property holds. Finally, if we are in the case where the two reachable sets overlap only on one side, i.e., a situation in which $(\mathcal{R}_\ell < \mathcal{Y}_\ell \wedge \mathcal{R}_u \leq \mathcal{Y}_u)$ or the symmetric case, $(\mathcal{R}_\ell \geq \mathcal{Y}_\ell \wedge \mathcal{R}_u > \mathcal{Y}_u)$, we cannot state whether the property is respected or not, and we have to proceed with the iterative refinement process \cite{reluval}. This process is repeated until, in all the partitions of the input space created with the iterative refinement, we have complete safety or violation.

\begin{figure}[h!]
    
    \centering
    \includegraphics[width=0.8\linewidth]{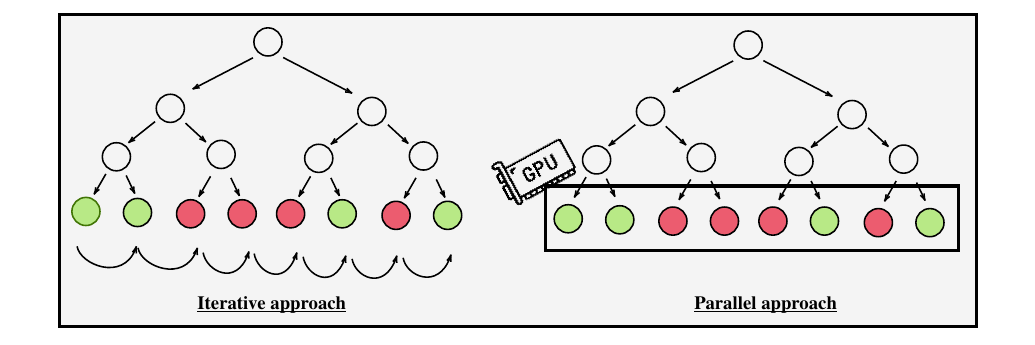}
    \caption{Left: the original approach proposed in \cite{CountingProVe} for obtain the exact count for the \textit{\#DNN-Verification} problem. Each node of the BaB is explored iteratively, thus resulting in poor sample efficiency and scalability as the size of the tree grows. Right: part of the optimization proposed in this work. As shown in \cite{ProVe, BetaCrown}, we can exploit GPUs to verify each layer of the BaB in parallel, thus enhancing the performance of exact count tools.}
    \label{fig:parallel}
\end{figure}

Even though the application of SLR improves the speed up of the verification process, the main difference with the \textit{DNN-Verification} problem is that we have to check every single node of the BaB tree. Hence, an iterative approach like the one represented in the left part of Fig. \ref{fig:parallel}, can be very inefficient as the number of branches increases. In contrast, the idea is to fully exploit the power of modern accelerators, i.e., the GPUs, to compute in a parallel fashion the result of the FV verification of each layer of the BaB tree, as shown in \cite{ProVe, BetaCrown} and reported in the right Fig. \ref{fig:parallel}. 
In particular, each node located in the same layer $t$ of the BaB tree represents a separate sub-instance to analyze and thus can be threaded independently. We represent the total number of sub-instances at the same layer $t$ with a value $n$ and the cardinality of each sub-instance with $k$. Thus, each layer $t$ of the BaB tree can be represented using a matrix of size $(n, k, 2)$, where the number $2$ represents the lower and upper bound of each of the $n$ intervals of cardinality $k$ to be verified. To clarify, referring to Fig.\ref{fig:parallel} and assuming we want to parallel verify the last level of the BaB tree shown on the right side of the image, this would be encoded with a matrix $(8, k, 2)$, with $k$ the dimension of the input space.

The main difficulties in implementing SLR on GPUs in CUDA are related to the management of memory cost. When comparing SLR with naive propagation, we can notice that the propagation of the coefficients of two equations for SLR, as compared to the propagation of the upper and lower bound values for naive, requires more memory in proportion to the number of input nodes in the network.
In fact, given $m$ the maximum layer size and $k$ the number of input nodes, naive propagation requires $m \cdot 2$ floats in memory to propagate the upper and lower bound on each node in the layer, while SLR requires $m \cdot k \cdot 2$ floats in memory to propagate the coefficients associated with each input node through the DNN. Since both the global and shared memory of each block in the GPU is generally limited, performing the verification on DNNs with either a considerable number of input nodes or particularly large hidden layers can be challenging on GPUs. 
To address this issue, we implement a parallel version of symbolic linear relaxation via a CUDA kernel, which concurrently runs interval analysis on each of the $n$ sub-instances by assigning each one to a separate CUDA thread. In order to further improve the computational efficiency, we load the lower and upper equations for symbolic linear relaxation onto the shared memory, which provides faster access time compared to the global memory, vectorizing every data structure into one-dimensional arrays. Moreover, to address the memory limitation discussed above, we maintain a small number of threads per block within limits allowed by the ratio of shared memory and equation size for the network we want to verify.
The resulting output bounds for each of the $n$ sub-instances verified are then saved on the specific indices of an output vector and analyzed to compute the next layer $t+1$ of the BaB tree.

% The numpy array is vectorized, each input subinterval is assigned to its corresponding CUDA thread via indexing and the resulting output bounds are saved on the appropriate indices of an output array with size $(n, m, 2)$, where $m$ is the number of output nodes. The lower and upper equations for symbolic linear relaxation are implemented as a single array with size $max\_layer\_size * (2 * input\_size + 2)$, where $max\_layer\_size$ is the number of nodes in the largest layer of the network and $input\_size$ is the number of input nodes. In order to further improve the computational time, the equations are loaded onto the shared memory, which provides faster access time compared to the global memory, and every data structure is vectorized into one-dimensional arrays.

In the next section, we see how the combination of a fast, tight parallel propagation of bounds, and thus, formal verification process, allows us to speed up the existing \textit{\#DNN-Verification} tools \cite{ProVe, CountingProVe}.
\section{Empirical Evaluation}

In this section, we empirically analyze the improvement in scalability and efficiency of the methods proposed in this work. In particular, our goal is to show that the combination of efficient techniques employed for standard FV can be extended for the \textit{\#DNN-Verification}, enhancing the computational efficiency of existing tools and thus allowing the adoption of these methodologies also in realistic robotic applications.

To this end, we applied the optimization discussed in Sec. \ref{method} on \texttt{ProVe}\cite{ProVe} and we refer to this new method as \texttt{ProVe\_SLR} and \texttt{CountingProVe}\cite{CountingProVe} referred as \texttt{CountingProVe\_SLR}. Moreover, we compare our methods against the \texttt{Exact Count} presented in Alg. \ref{alg:exact}, which use naive interval propagation and no parallelization. All the data have been collected on a commercial PC running Ubuntu 22.04 LTS equipped with Intel i5-13600KF and Nvidia GeForce RTX 4070 Ti, reporting for each method the violation rate and the computation time.

We show the results of our experiments on three different case studies.  
In the first block of Tab. \ref{tab:results}, we analyze ACAS Xu\cite{ACAS, Reluplex}, a realistic safety-critical domain in which finding all possible unsafe configurations is crucial. In more detail, ACAS Xu is an airborne collision avoidance system for aircraft, a well-known benchmark for FV of DNNs. It consists of 45 models with five input nodes, five output nodes, and six hidden layers containing 50 $ReLU$ nodes each. In our experiments, we compare the performance of each method on three different models for which we have a \texttt{SAT} answer (i.e., at least a violation point) on the property $\phi_2$, which describes a scenario where, if the intruder is distant and significantly slower than the ownship, the score of a Clear of Conflict (CoC) can never be maximal\footnote{We refer the interested reader to the original paper \cite{ACAS, Reluplex} for more details on this property.}.

\begin{figure}[h!]
    %\vspace{-3mm}
    \centering
    \includegraphics[width=0.7\linewidth]{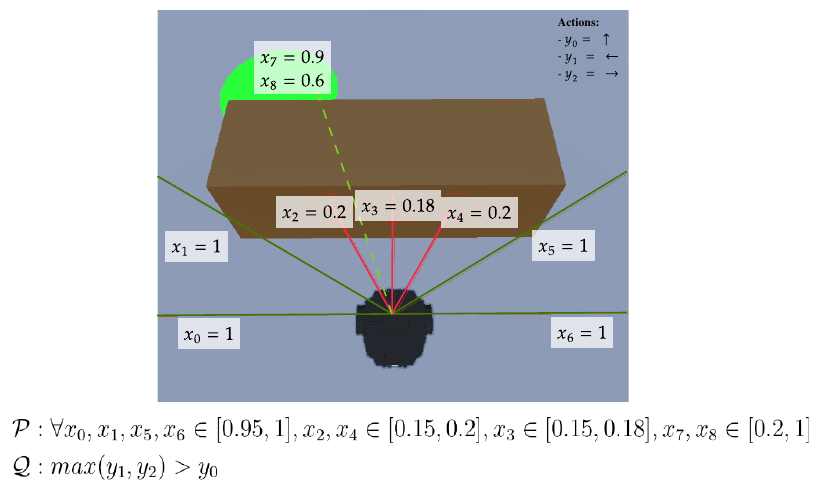}
    \caption{Explanatory image of the safety property tested on different models in the second block of Tab \ref{tab:results}. Each input $x_0,\dots,x_6$ represents a lidar scan value between $[0,1]$, while $x_7,x_8$ are heading and distance from the goal, respectively. Hence, the safety property ensures that in this situation, a forward movement ($y_0$) is not chosen by the agent.}
    \label{fig:property}
\end{figure}

We then analyze a set of DRL Mapless Navigation models collected in recent works \cite{TACAS, aamas_safety}. In particular, in the second block of Tab.\ref{tab:results}, we consider an input space composed of nine nodes, two hidden layers of 16 nodes activated with ReLU, and three output nodes that encode discrete actions that the agent can perform in the environment. We chose discrete actions over continuous ones as discrete controllers can achieve excellent performance in robotic navigation even in multi-agent settings \cite{multiagent1,multiagent2}.

Finally, in order to test the scalability, in the last block of Tab. \ref{tab:results}, we increase the complexity and test a model with 13 inputs representing more dense laser scans, two hidden layers of 64 ReLU nodes, and six output nodes that encode, once again, discrete actions for the robot. The safety properties tested in both these mapless navigation scenarios are behavioral properties, as the one described in Sec.\ref{DNN-ver}. We provide the reader with a pictorial representation of the property tested in Fig. \ref{fig:property}.

Focusing on the first block in Tab. \ref{tab:results}, due to the scalability issues discussed in Sec.\ref{method}, the \texttt{Exact Count} method reaches the timeout (fixed after 24 hours), failing to return an exact answer for all the tests performed. 
\begin{table}[t]
\scriptsize
\begin{tabular}{|cc||cc||cc||cc||cc|}
\hline  
\multicolumn{2}{|l||}{} &
  \multicolumn{4}{c||}{\textbf{Exact methods}} &
  \multicolumn{4}{c|}{\textbf{Approximations (confidence \textgreater 99\%)}} \\ \hline
\multicolumn{2}{|c||}{\textbf{Instance}} &
  \multicolumn{2}{c||}{\texttt{Exact Count}} &
  \multicolumn{2}{c||}{\texttt{ProVe\_SLR}} &
  \multicolumn{2}{c||}{\texttt{CountingProVe}} &
  \multicolumn{2}{c|}{\texttt{CountingProVe\_SLR}} \\
\multicolumn{2}{|c||}{}          & VR & Time & VR & Time &  Lower Bound VR & Time & Lower Bound VR & Time \\ \hline
\multicolumn{2}{|c||}{$\phi_2$ ACAS Xu\_4.3}      & \xmark   &  $>$ 24h    &  1.43\%   &  \textbf{8h46m}    &  1.32\%  &  2h4m    &  1.28\%  &   \textbf{45m}   \\
\multicolumn{2}{|c||}{$\phi_2$ ACAS Xu\_4.9}      &  \xmark   &  $>$ 24h      &  0.15\%  & \textbf{12h21m}     & 0.093\%   & 2h23m     &  0.10\%   & \textbf{59m}    \\
\multicolumn{2}{|c||}{$\phi_2$ ACAS Xu\_5.8}      &    \xmark   &  $>$ 24h     &  2.2\%  & \textbf{4h35m}     & 1.96\%   &  2h10m    &   1.76\% &  \textbf{51m}    \\ \Xhline{2.5\arrayrulewidth}
\multicolumn{2}{|c||}{Model\_9\_1}  &    \xmark   &  $>$ 24h      &  27.98\%  & \textbf{3h46m}     &  26.47\%  &    2h17m  &   25.98\% & \textbf{42m}     \\
\multicolumn{2}{|c||}{Model\_9\_2}  &    \xmark   &  $>$ 24h      &  20.88\%  & \textbf{4h20m}     &  19.73\%  & 2h31m     & 19.95\%   &  \textbf{48m}    \\
\multicolumn{2}{|c||}{Model\_9\_3}  &    \xmark   &  $>$ 24h      &  25.62\%  &   \textbf{3h12m}   &  24.64\%  &  2h2m    &  24.26\%  & \textbf{47m}     \\ \Xhline{2.5\arrayrulewidth}
\multicolumn{2}{|c||}{Model\_13\_64} &    \xmark   &  $>$ 24h      &  22.13\%  & \textbf{5h41m}     &  20.99\%  &  3h21m    & 21.24\%   & \textbf{1h30m}     \\\hline
%\multicolumn{2}{|c||}{Model\_MN\_2} &    \xmark   &  $>$ 24h      &    &      &    &      &    &      \\
%\multicolumn{2}{|c||}{Model\_MN\_3} &    \xmark   &  $>$ 24h      &    &      &    &      &    &      \\ \hline
\end{tabular}
\caption{Comparison on different case studies of \texttt{Exact Count} without parallelization, \texttt{ProVe\_SLR} that combines parallelization and SLR, CountingProVe of \cite{CountingProVe} and \texttt{CountingProVe\_SLR} which exploits the efficiency improvements of \texttt{ProVe\_SLR} as backend. The first block shows the results on the Acas Xu $\phi_2$ FV benchmark. The second and third blocks report the results of the robotic mapless navigation scenario.}
\label{tab:results}
\end{table}
On the other hand, \texttt{ProVe\_SLR} is able to return the exact VR in about 8 hours and 34 minutes (mean computation time), which corresponds to a $\sim 64.3\%$ mean time reduction to compute the VR rate. The time required by \texttt{ProVe\_SLR} is still considerable, primarily due to the need to write partial results to disk during the verification process to address memory constraints. We specify that with more powerful hardware, the verification times can be improved further, but we argue that it is interesting to show how the optimizations proposed in this work allow the use of these verifiers in safety-critical robotic scenarios, even using commercial hardware. 

A slight minor improvement is also obtained when we use the \texttt{CountingProVe\_SLR} approximation that uses \texttt{ProVe\_SLR} as backend over the original \texttt{CountingProVe}\cite{CountingProVe}. Recalling the intuitions of this latter approximation provided in Sec. \ref{background_sharpDNN}, given the improvement in the scalability of \texttt{ProVe\_SLR}, we are able to reduce the number of splits required before employing the exact count on the leaf, thus reducing the computational time required to compute the lower bound of the VR. In particular, focusing on the results, given the randomized nature of \texttt{CountingProVe}, the VR reported by this latter and \texttt{CountingProVe\_SLR} is slightly different. Nonetheless, both the results are correct as we obtain a lower bound of the real violation rate computed by \texttt{ProVe\_SLR}. The significant difference in this first block is that we obtained a $\sim$ 61.4\% mean reduction in the computation time between \texttt{CountingProVe} and \texttt{CountingProVe\_SLR}.

Regarding the experiments of robotic mapless navigation, also in this realistic scenario, we notice a substantial improvement in computation time when applying the combination of SLR and parallel computing. Crucially, in both these sets of experiments, we obtain a mean computational improvement of $\sim 80\%$ when using \texttt{ProVe\_SLR} over a naive exact count approach and a $\sim 61.6\%$ when employing \texttt{CountingProVe\_SLR} over \texttt{CountingProVe}.

\subsection{Limitation of Sampling Strategies}

As discussed in Sec. \ref{background_sharpDNN}, if we are interested in an estimation of the violation rate, there are sampling-based techniques with probabilistic guarantees that allow us to obtain a reasonably accurate result, significantly reducing the computation time by orders of magnitude. However, we argue that in safety-critical contexts such as robotic DRL, where expensive hardware and human lives are potentially at risk, having an estimate of the number of unsafe input configurations may not be sufficient. To this end, we performed an additional test considering a simple DNN with three input nodes, two hidden layers of 32 ReLU nodes, and a single output. We tested the following property:
 \begin{align*}
     \mathcal{P}&: \forall x_0,x_1,x_2 \in [0, 1]\\
     \mathcal{Q}&: y \geq 0 
 \end{align*}
Hence, we collected an estimate of the violation rate for this network and this property by employing a Monte Carlo sampling approach using $1M$ samples. The result obtained with this strategy reported a VR = 0\%, i.e., no violation points in the input domain. We then perform the formal verification on the same DNN and property, employing \texttt{ProVe\_SLR}, obtaining instead a VR = 0.00012\%. As previously discussed, this result shows that sampling-based approaches are computationally more efficient; still, in the presence of low VR rates, they can fail even with considerable sample sizes. Hence, it is crucial to develop efficient and very accurate approximation methods capable of scaling on larger instances of the problem, capturing even the minute fraction of the violation rate potentially present in the property's domain.

\section{Discussion}

In this paper, we present efficient optimizations to the existing solutions for the \textit{\#DNN-Verification} problem. Crucially, we show that the combination of symbolic linear relaxation for tight output reachability sets estimation and parallel computation on GPUs enhances the scalability and performance of existing \textit{\#DNN-Verification} tools. This work paves the way for the application of different types of FV in very complex, realistic robotic scenarios, such as mapless navigation. We show how naive sample-based approximations can fail to capture a minimal fraction of the violation rate in the property's domain, emphasizing the necessity for more sophisticated approximation methods for counting violation points such as \texttt{CountingProVe\_SLR}.

Future directions involve developing novel approximation algorithms capable of scaling to considerably larger instances of the problem, being able to provide the violation rate even in complex contexts such as robotic DRL. Moreover, we plan to extend the optimizations proposed in this work to different robotic tasks, even with different input types, such as images.

%% Define the bibliography file to be used
\bibliography{sample-ceur}

\end{document}